\documentclass{article}

\usepackage[utf8]{inputenc}

\usepackage[final]{mlrsa_2020}
\usepackage[utf8]{inputenc} 
\usepackage[T1]{fontenc}    
\usepackage{hyperref}       
\usepackage{url}            
\usepackage{booktabs}       
\usepackage{amsfonts}       
\usepackage{nicefrac}       
\usepackage{microtype}      
\usepackage{graphicx}

\title{Survey on Modeling Intensity Function of Hawkes Process Using Neural Models}

\author{%
  Jayesh Malaviya \\
  Computer Science and Engineering\\
  IIT Gandhinagar\\
  \\
  \texttt{malaviya$\_$jayesh@iitgn.ac.in} \\
}

\begin{document}

\maketitle

\begin{abstract}
  The event sequence of many diverse systems is represented as a sequence of discrete events in a continuous space. Examples of such an event sequence are earthquake aftershock events, financial transactions, e-commerce transactions, social network activity of a user, and the user's web search pattern. Finding such an intricate pattern helps discover which event will occur in the future and when it will occur. A Hawkes process is a mathematical tool used for modeling such time series discrete
events. Traditionally, the Hawkes process uses a critical component for modeling data as an intensity function with a parameterized kernel function. The Hawkes process's intensity function involves two components: the
background intensity and the effect of events' history. However, such parameterized assumption can not capture future event characteristics using past events data precisely due to bias in modeling kernel function.
This paper explores the recent advancement using novel deep learning-based methods to model kernel function to remove such parametrized kernel function. In the end, we will give potential future research directions to improve modeling using the Hawkes process.

\end{abstract}

\section{Introduction}

Collections of random data points in some space, such as time or location modeled as point processes. The point processes give the statistical language for describing the timing and properties of the events. We can fit many problems in this set of point processes with a range of application domains. Like in finance, a process can represent a sell or a buy transaction in the stock market that influences future prices. In geophysics, the process can be an earthquake that is implying the likelihood of a future earthquake in the vicinity. In social media analytics, the process can be over a period of a time user action, each of that has a number of properties like user influence, connectivity with the neighboring network, and the topic of interest in the community. Hawkes processes are one of the point processes, and especially useful for modeling discrete, inter-dependent events over continuous time [1,2].

Some times we can directly observe the diffusion of information via retweets in case of twitter, but there exist other diffusion processes that can not be observed clearly like offline word-of-mouth diffusion, or propagating of information over emails and different online platforms. One way for modeling such an information diffusion process is by self-exciting processes. In these processes, we are seeing the probability of seeing a new event increases due to previous events.

Many events are by default correlated in the surrounding, viz. a single or multiple events can cause or stop the future event from occurring. Papers in this domain are mainly interested in learning the distribution of such time-series events and various events' properties. If we can concretely model the correlation among these events, we can predict the future event or set of events using the history of future events' events and timing.  An example of such correlation among events can be found in many applied domains. Every patient has a sequence of a doctors visit, laboratory tests, diagnoses, and medications in medical events. By learning from past patients sequence data, we can predict a new patient's future events from their past events. In e-commerce, a user has a sequence of online purchase transactions, and by modeling such a sequence, we can learn the consumers' purchase patterns. In social network data, the user actions are previous posts, likes, comments, and messages, and from that, we can predict future activities of the users [1, 4].

This paper's primary motivation is despite being much traditional literature on time series data modeling using the Hawkes process using many different kernel functions empirically tested for specific applications. Still, they can't generalize well and suffer from the problem of the curse of model misspecification. To tackle that, recently, people started exploring deep learning-based kernel function modeling techniques to avoid many parametric assumptions about intensity function and showing useful improvements in the future event prediction and efficiently handling high dimensional time series compared to traditional approaches.

\subsection{Motivation for the survey}

Hawkes process is traditionally used to predict future events or event causality detection. These two closely match machine learning objectives, viz. models prediction accuracy and interpretability [Choi et al., 2016]. Traditionally till recently, people were using a manually specified parametric kernel function to model Hawkes processes. Recently, people started exploring deep neural models for modeling the Hawkes process's intensity function to remove the curse of model misspecification. More recent neural models call it neural Hawkes process as the term used in [Du et al., 2016; Mei and Eisner, 2017], which removes the need for explicit parametric intensity function.  Du et al. also showed that the classical Hawkes process models with parametric intensity function are good at interpretability on understanding the learning process and event causality.

In contrast, the neural Hawkes process gives high accuracy in predicting future events and time. In the subsequent section, we will review the more recent work along these threads. As this is still an emerging area, there are fewer papers available for the neural Hawkes process. Nevertheless, current curiosity in top Machine Learning conferences like ICML, Neurips, etc. about this topic shows a great scope of improvement in this line by leveraging deep learning techniques to predict future events and interpretability of causal relationships better.

\section{Preliminary}
In this section we will describe basic definitions and terminology used in modeling Hawkes process.

\paragraph{Definition 1 (Point Process [1])} Let $(\omega, F , P)$ be a probability space. A point process on $R^{+}$ is a sequence of non-negative random variables $\left \{ t_i \right \}_{i=1,2, \dots}$ , such that for all i, $t_i \leq t_{i+1}$.

\paragraph{Definition 2 (Interarrival time [1])} Consider a temporal point process $\left \{ t_i \right \}_{i=1,2, \dots}$ , Define $W_i = t_i - t_{i-1}$ with the convention $t_0 = 0$. The sequence $\left \{ W_i \right \}_{i=1,2, \dots}$ is called the interarrival times.

\paragraph{Definition 3 (Counting Process [1])} A counting process associated with the temporal point process $\left \{ t_i \right \}_{i=1,2, \dots}$ represented as $N_t$ or $N (t)$ is a random function defined on time $t\geq 0$ ,  taking values in $Z^{+}$ , that counts the number of occurrences up to and including time $t$, i.e. 

$$N_t = \sum _{i= 1,2, \dots} 1_{t_i \leq t}$$

\paragraph{Definition 4 (Intensity of point process [1])} The intensity of a point process is defined as,

$$\lambda (t) = \lim_{h -> 0^{+}} \frac{P(N(t, t+h] > 0)}{h}$$

$N(t, t+h]$ indicate the count of the number of events in an interval of length h. $\lambda (t)$ intuitively measures the rate at which the events occur.

A fundamental model for modeling sequence data is the Poisson process, which assumes that the events arriving independently of other events. In a typical homogeneous Poisson process, we take that the intensity function is constant $\lambda (t) = \lambda$
. However, in a non-homogeneous Poisson process, event probability t varies with time t. Still, it is independent of another process. In a Hawkes process (Hawkes, 1971; Liniger, 2009), they incorporated that the past events can temporarily raise future event probability. They assume that this self-excitation is positive, ­ additive over the past events, and exponentially decaying with time. 

\paragraph{Definition 5 (Hawkes process [2, 1])} Let $\left \{ N_t \right \}_{t>0}$ be a counting process with associated history $H_t , t \geq 0$. The point process is said to be a Hawkes process if the conditional intensity function $\lambda (t|H_t)$ takes the form,

$$ \lambda (t|H_t) = \lambda_0 (t) +  \sum _{i: t > T_i } \phi (t- T_i)$$ 

where $T_i<t$ are all the event time having occurred before current time t, and which contribute to the event
intensity at time t. $\lambda_0 (t) : R \mapsto R_+$ is a deterministic base intensity function, and $\phi :  R \mapsto R_+$  is called the
memory kernel, normally kernel is stocastic and depend on the past events throuth the kernel function.

\section{Traditional Hawkes process likelihood and Kernel functions}

The process is described via a conditional intensity
function $\lambda (t|H_t)$ [1, 6], which is intensity function at time t conditioned on history of event $H_t = \left \{ t_i \right \}_{t_i < t}$ upto time t, 

$$\lambda (t|H_t) = \lim_{\Delta \mapsto 0} \frac{P(one \: event \: occur \: in [t, t+\Delta) | H_t )}{\Delta}$$

If conditional intensity function is specified then, the probability density function of the next event time $t+1$, given the past event times $\left \{ t_1, t_2, \dots t_i\right \}$  is [1, 6] calculated as,

$$P(t_{i+1} | t_1 , t_2, \dots , t_i) = \lambda (t_{i+1} | H_{i+1}) \, exp \left \{ - \int_{t_i}^{t_{i+1}} \lambda (t| H_t) dt \right \}$$

where the exponential term in the right-hand side represents the probability that no events occur in
$[t_i; t_{i+1})$. The probability density function to observe an event sequence $\left \{t_i\right \}_{i=1}^{n}$ is [1, 6] calculated as,

$$P(\left \{ t_i \right \}_{i=1}^{n}) = \prod_{i=1}^{n} \lambda (t_{i} | H_{t_i}) \, exp \left \{ - \int_{0}^{T} \lambda (t| H_t) dt \right \}$$

Detailed proof for the above equations is available in [1].

\paragraph{Conditional intensity functions of point processes} 

\begin{itemize}

\item Poisson Process (Kingman 1992): As discussed above, the Poisson process has a constant intensity function and no event history effect on current events.

\item Reinforced Poisson processes (Pemantle 2007; Shen et al. 2014): the model captures the 'rich-get-richer' phenomenon defined by a compact intensity function, which
is lately used for predicting popularity (Shen et al. 2014).

\item Hawkes process (Hawkes 1971): Hawkes process incorporates constant background intensity with kernel function for modeling the event's history, and it is a self-excitation process. It received wide recognition lately in analysing the social network data
(Zhou, Zha, and Song 2013a), viral diffusion(Yang and
Zha 2013) etc.

\item Reactive point process (Ertekin, Rudin, and Mc-
Cormick 2015): It can be thought of as a generalization of the Hawkes process. It generally combines a self-inhibiting term in the Hawkes process to account for inhibiting effects from past events.

\item Self-correcting process (Isham and Westcott 1979): In this background, part increases steadily, while it is reduced by a constant $e^{- \alpha} < 1$ every time a new event appears.

\end{itemize}

Above is table 1 of the conditional intensity functions of various point processes in their general form. In that, it is shown background intensity component and event history effect component separately.

\begin{table}
  \caption{Conditional intensity functions of point processes \ [4]}
  \label{Conditional-intensity}
  \centering
  \begin{tabular}{lll}
    \toprule
    
    Model     & Background     & History event effect \\
    \midrule
    Poisson process            &  $\mu (t)$          & 0                     \\
Reinforced poisson process &     0       &   $ \gamma (t) \sum _{t_i < t} \delta (t_i - t)$                  \\
Hawkes process             &    $\mu (t)$        &   $\sum _{t_i < t} \gamma (t,t_i)$                   \\
Reactive point process     &   $ \mu (t) $       &   $\sum _{t_i < t} \gamma_1 (t,t_i) - \sum _{t_i < t} \gamma_2 (t,t_i)$                   \\
Self-correcting process    &    0        &  $exp(\mu t - \sum _{t_i < t} \gamma (t,t_i))$                    \\ 
    \bottomrule
    
  \end{tabular}
   
\end{table}

Note: $\delta (t)$ is Dirac function, $\gamma (t, t_i)$ is time-decaying kernel and $\mu (t)$ can be constant or time-varying function.

\subsection{Problems}

The real-world pattern can not correctly model using the Hawkes process using the assumption for intensity function like assumption about function is positive is violated if a current event inhibits future events rather than excite. For example, Softdrink consumption inhibits alcohol consumption. Another assumption of additivity is infringed in past events that are not additive, i.e., the 100th advertisement may not help much compared to the first advertisement to increase the purchase rate and even decrease the purchase. The third assumption of exponential decrease with respect to time is violated when a prior event has a very delayed effect, so it may increase sharply just before decaying. 

Another critical problem with the classical Hawkes process is complexity as to generalize the model well, we usually pick complex kernel function, which will lead to the intractable likelihood function due to the integral of a kernel function. However, many numerical approximation algorithms are available to compute integral but generally deteriorate the model's accuracy and computationally very difficult to train for high dimensional time series.

Another problem is missing data in the observed sequences, which generally requires a more robust model than the traditional Hawkes process.  Manier time, even if a particular domain Hawkes model might be appropriate, it is still hard to apply when sequences of events are partially observed. Real datasets sometimes systematically exclude some types of events, i.e., offline purchases, which may influence future events.

\section{Neural Models for Hawkes process}

After investing in different forms of conditional intensity functions[11, 12, 13], the authors in [3] observed that different forms of dependency structures exist among past events and future events. With this crucial insight, they found that to learn general representation for approximating the latent dependency among the event's history is essential for predicting future event time and the type of event.

In subsequent sections, we will explore various deep learning techniques to model the Hawkes process's conditional intensity function. In Table 2, Neural Model-Based Methods with Implementation Code are provided for exploration purposes.

\subsection{RMTPP: Embedding Event History to Vector [3]}

A recurrent neural network (RNN) with sigmoid activation units can capable of simulating a universal Turing machine [22], which can use to perform complex computations. RNN has shown promising results in the sequence modeling task. For example, in NLP, the recurrent neural network has an excellent result for predicting future sequence and sequence to sequence translations[23]. It has been used in discrete time series data prediction[24, 25, 26] for a long time. 

In this paper, the author proposes the Recurrent marked temporal point process to the model type of the event, which is event marker and event timing simultaneously. Their approach's main idea is to use RNN in place of modeling intensity function by assuming it a complex non-linear function; an RNN can automatically learn the relationship between past events and future events and their influences. RMTPP is the first attempt to incorporate a deep model into the temporal point process, and results show improvement over traditional methods. RMTPP can also scale to millions of events and get better predictions compared to parameterized methods.

Input for the RNN in this approach is the occurrence timing of the event $t_j$ and the type of the event $y_j$, the pair $(t_j, y_j)$ 
and then RNN unrolled up to the event $(j+1)^{th}$ event. The embedding $h_{j-1}$ represents the past events influence memory where both timing and type of event are captured.  The neural model updates $h_{j-1}$ to $h_j$  by considering the effect of the event $(t_j, y_j)$.  With this, we can now represent the conditional intensity of $j^{th}$ event as a $h_j$ since it represents the history of influence till $j^{th}$ event,

$$f^*(t_{j+1}) = f(t_{j+1}|H_t) = f(t_{j+1}|h_j) = f(d_{j+1}|h_j)$$

where $d_{j+1} = t_{j+1} - t_{j}$. So now we can predict next events timing $\hat{t}_{j+1} $ and event type $\hat{y}_{j+1} $, that is depend on the $h_j$. 

The advantage of this method is it can now capture a hidden representation of event history in $h_j$ that captures a more general form of event representations concerning time without being relay on the parameterized assumption in case of intensity function $\lambda (t|H_t)$.

The disadvantage of this approach is that we lose model interpretation and causal relationships among the events.

\subsection{NHP: A Neurally
Self-Modulating Multivariate Point Process [5]}

In this paper author modeled a sequence of discrete events in continuous time by a novel ideal of continuous-time LSTM. Like before, each event type k has its intensity $\lambda_k (t)$, which varies with time and jumps discontinuously when each new event occurs and, in the end, drifts continuously towards baseline intensity. In this continuous-time LSTM, we have hidden state vector $h(t) \in (-1,1)^D$, which depends on a vector $c(t) \in R^D$ of the memory cell. This model is different from traditional discrete-time LSTM(Hochreiter and Schmidhuber, 1997; Graves, 2012) in the continuous interval following an event, and each memory cell c exponentially decays at some rate $\delta$ towards some steady-state value $\bar{c}$, 

At each time t > 0, we obtain the intensity $\lambda_k (t)$ by $\lambda_k (t) = f_k(W_k^t  h(t))$, where $h(t) = o_i \odot (2\sigma (2c(t)) - 1)$ for $t \in (t_{i-1}, t_i]$, defines how the hidden states $h (t)$ are continually obtained from the memory cells $c(t)$ as the cells decay. This basically says that on the interval $ (t_{i-1}, t_i]$, after occurrence of $i-1$ event till $i^{th}$ event occur at time $t_i$ , the $h(t)$ define in above equation determines the intensity function $\lambda_k (t) = f_k(W_k^t  h(t))$.

This paper is different from the previous on RMTPP because the RMTPP uses standard discrete-time LSTM cell without decaying like in NHP, so they encode intervals between past events in terms of numerical input the LSTM. Another is they use only one intensity function $\lambda (t)$, and it at the end, it merely decay towards 0 exponentially between two events, whereas in NHP, more modular mode creates separate intensity function $ \lambda_k (t)$. They each allow complex and non-monotonic dynamics towards non zero steady-state intensity. Another main difference is in RMTPP. They assume inherently that event type y  and event time t are conditionally independent given history h, and they are finding different distribution for them, and their model can not avoid the positive probability of extermination at all times. Lastly, they taking exponential function, and therefore the effect of their hidden unit of RNN is multiplicative, whereas in this paper, the function is softplus, so here they are getting additive effect inspired by classical Hawkes process. As per the author in NHP, additivity helps capture independent causes, and at the same time, hidden units in NHP can capture complex joint causes.

\subsection{Fully Neural Network based Model for General
Temporal Point Processes [6]}

In the past literature[3, 5], RNN based models usually assume some specific functional forms for the time course of the intensity function of a temporal point process, i.e., exponentially increasing or decreasing with time since most recent events. However, such an assumption limits the expressibility of the model. So in this paper, the author proposes an exciting ideal of RNN based model. They remove the assumption of the time course of intensity function and represent it as a generic way with a cumulative hazard function. In this paper, the first model intensity functions integral using a feedforward neural network and then obtain it as a derivative. With this approach, we can evaluate the exact log-likelihood function, which otherwise we have to use a numerical approximation algorithm and computationally expensive.

\subsection{Self-Attentive Hawkes Process [7]}

In this paper, the author examines the use of the self-attention mechanism in the Hawkes process by proposing a Self-Attentive Hawkes Process (SAHP). First, they apply self-attention to estimate the influence of the history of events on the next event by computing its probability. As self-attention relies on positional embeddings of the input for taking into account the events' order, traditional embedding methods are used on sinusoidal functions for shifting their position by a constant shifting phase, which if used for sequences would ignore the original time interval between events.  
This paper removes this limitation by proposing a novel idea of a time-shifted positional embedding method where they assume time interval as a phase shift of sinusoidal functions. This attention-based model is more interpretable compared to previous RNN based models as the learned attention weight vector can be used to reveal 
the contributions of one type of event in another kind of event happening and their causal relationships.

\subsection{Transformer Hawkes Process [8]}

Sequential event data often contain complicated long-term and short-term temporal dependencies. Previously studied literature on neural models with the Hawkes process fails to simultaneously capture both these types of dependencies and eventually give an unreliable prediction on future events and timestamps.

\begin{figure}[!ht]
 \centering
  \includegraphics[width=3in]{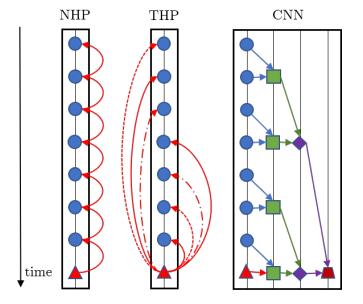}
  \caption{Illustration of dependency computation between the last event (the red triangle) and its history (the blue circles). RNN based NHP models dependencies through recursion. THP directly
and adaptively models the event’s dependencies on its history. Convolution-based models enforce static dependency patterns [8].}
\end{figure}

 In this paper author target this issue with the help of the Transformer Hawkes Process (THP) model, which uses the self-attention mechanism for capturing the long-term dependencies and, at the same time, computationally very efficient compared to previous models. Experiments of the paper on various datasets suggest that THP gives better likelihood and event prediction than existing methods by a good margin.

The main component of the THP model is the self-attention module, which, unlike RNN, discards recurrent structures (figure 1). However, this model still needs to be aware of the temporal information of input, which is the event time. Here, they are using trigonometric functions to define a temporal encoding for each timestamp $t_j$. Here, the self-attention mechanism model can directly select events whose occurrence time is at any distance from the current time. Whereas in RNN based models encoding of event history is sequential via a hidden unit representation of the events, i.e., the state $t_j$ depends on state $t_{j-1}$, which in turn depends on $t_{t-2} $ and so on. This encoding is weak, and RNN may fail to learn relevant information for any random event $t_k$, hidden unit for any event $t_j$ where $j \geq k$ will be inferior.

The limitation of previous RNN based models (figure 1) is the need to feed data sequentially, and because of that, it can not train parallelly. This limits RNN to train for large scale data. Recently CNN based models (Oord et al., 2016) [27] are also proposed to capture long term dependency in sequential data. But the limitation of that model is that these models are inherently enforced to capture many unnecessary dependencies.

They experimented with model sensitivity to the number of model parameters and tested on the Retweet dataset and found that THP is not sensitive to its number of parameters. The THP model outperforms RNN based models with very few parameters. In their experiments, they showed that NHP with about 1000k parameters matches the performance of the time-series event sequence (TSES) model with about 2000k parameters.  So THP is efficient in both model size and training time.

\begin{table}
\caption{Neural Model Based Methods with Implementation Code}

\begin{tabular}{p{0.50\linewidth}  p{0.50\linewidth}}
 
\hline

Neural Model Based Methods                                                    & Code                                           \\ \hline \\
RMTPP: Embedding Event History to Vector {[}3{]}                              & \url{https://github.com/musically-ut/tf_rmtpp}                \\
NHP: A Neurally Self-Modulating Multivariate Point Process {[}5{]}            &  \url{https://github.com/xiao03/nh}                            \\
Fully Neural Network based Model for General Temporal Point Processes {[}6{]} & \url{https://github.com/omitakahiro/NeuralNetworkPointProcess} \\

Self-Attentive Hawkes Process {[}7{]}                                         & \url{https://github.com/QiangAIResearcher/sahp_repo}   \\       
Transformer Hawkes Process {[}8{]}                                            & \url{https://github.com/SimiaoZuo/Transformer-Hawkes-Process}
\\ 
\hline

\end{tabular}
\end{table}

\section{Future Scope}

Deep learning-based conditional intensity modeling has opened a new research thread in the temporal point process domain. Still, at the same time, it compromises the interpretability of the model, and eventually, it makes it challenging to deal with the causality of events. The problem of predicting the distribution of events in case of missing sequence data or partially observed sequence still requires efforts to improve results. It is interesting to embed these neural models into reinforcement learning to discover the causal relationship among events and learn a global policy. Causality analysis, in the case of asynchronous events using an attention-based model, is interesting to explore. The joint distribution of the event marker and event time can be more informative than assume it to be independent.

\section*{Acknowledgement}

I thank Professor Anirban Dasgupta, Department of Computer Science and Engineering IIT Gandhinagar, for guidance on the Hawkes process topic and useful discussion. I would also like to thank Rachit Chhaya, Ph.D. Scholar at the Department of Computer Science and Engineering IIT Gandhinagar, for providing valuable feedback of the work during writing.

\section*{References}

\medskip

\small

[1] Rizoiu, Marian-Andrei, Young Lee, Swapnil Mishra, and Lexing Xie. "A tutorial on hawkes processes for events in social media." { \it arXiv preprint} arXiv:1708.06401 (2017).

[2] Hawkes, Alan G. "Spectra of some self-exciting and mutually exciting point processes." { \it Biometrika} 58, no. 1 (1971): 83-90.

[3] Du, Nan, Hanjun Dai, Rakshit Trivedi, Utkarsh Upadhyay, Manuel Gomez-Rodriguez, and Le Song. "Recurrent marked temporal point processes: Embedding event history to vector." { \it In Proceedings of the 22nd ACM SIGKDD International Conference on Knowledge Discovery and Data Mining}, pp. 1555-1564. 2016.

[4] Xiao, Shuai, Junchi Yan, Xiaokang Yang, Hongyuan Zha, and Stephen M. Chu. "Modeling the intensity function of point process via recurrent neural networkss." {\it In Proceedings of the Thirty-First AAAI Conference on Artificial Intelligence}, pp. 1597-1603. 2017.

[5] Mei, Hongyuan, and Jason M. Eisner. "The neural hawkes process: A neurally self-modulating multivariate point process." {\it In Advances in Neural Information Processing Systems}, pp. 6754-6764. 2017.

[6] Omi, Takahiro, and Kazuyuki Aihara. "Fully neural network based model for general temporal point processes." {\it In Advances in Neural Information Processing Systems}, pp. 2122-2132. 2019.

[7] Zhang, Qiang, Aldo Lipani, Omer Kirnap, and Emine Yilmaz. "Self-Attentive Hawkes Process." {\it ICML}, 2020.

[8] Zuo, Simiao, Haoming Jiang, Zichong Li, Tuo Zhao, and Hongyuan Zha. "Transformer Hawkes Process." {\it ICML} 2020.

[9] E. Choi, M. T. Bahadori, J. Sun, J. Kulas, A. Schuetz, and W. Stewart. Retain: An interpretable predictive model for healthcare using reverse time attention mechanism. {\it In NIPS}, pages 3504–3512, 2016.

[10] A.G Hawkes. Hawkes processes and their applications to finance: a review. {\it Quantitative Finance}, 18(2):193–198, 2018.

[11] K. Cho, B. van Merrienboer, D. Bahdanau, and Y. Bengio.
On the properties of neural machine translation:
Encoder-decoder approaches. {\it CoRR}, abs/1409.1259, 2014.

[12] D. Daley and D. Vere-Jones. An introduction to the theory
of point processes: volume II {\it general theory and structure}, Springer, 2007.

[13] N. Du, M. Farajtabar, A. Ahmed, A. J. Smola, and L. Song.
Dirichlet-hawkes processes with applications to clustering
continuous-time document streams. {\it In KDD} , ACM, 2015.

[14] Kingman, J. F. C., Poisson processes. volume
3. {\it Oxford university press}, 1992.

[15] Pemantle, Robin. "A survey of random processes with reinforcement." {\it Probability surveys} 4 (2007): 1-79.

[16] Shen, Hua-Wei, Dashun Wang, Chaoming Song, and Albert-László Barabási. "Modeling and predicting popularity dynamics via reinforced poisson processes.", 
{\it In AAAI} 2014.

[17] Zhou, Ke, Hongyuan Zha, and Le Song. "Learning triggering kernels for multi-dimensional hawkes processes."{ \it In International Conference on Machine Learning}, pp. 1301-1309. 2013.

[18] Zhou, Ke, Hongyuan Zha, and Le Song. "Learning social infectivity in sparse low-rank networks using multi-dimensional hawkes processes." {\it In Artificial Intelligence and Statistics}, pp. 641-649. 2013.

[19] Yang, Shuang-Hong, and Hongyuan Zha. "Mixture of mutually exciting processes for viral diffusion." {\it In International Conference on Machine Learning}, pp. 1-9. 2013.

[20] Ertekin, Şeyda, Cynthia Rudin, and Tyler H. McCormick. "Reactive point processes: A new approach to predicting power failures in underground electrical systems." {\it The Annals of Applied Statistics} 9, no. 1 (2015): 122-144.

[21] Isham, Valerie, and Mark Westcott. "A self-correcting point process." {\it Stochastic processes and their applications} 8, no. 3 (1979): 335-347.

[22] H. T. Siegelmann and E. D. Sontag. Turing computability
with neural nets. {\it Applied Mathematics Letters}, 4:77–80,
1991.

[23] Sutskever, Ilya, Oriol Vinyals, and Quoc V. Le. "Sequence to sequence learning with neural networks." {\it In Advances in neural information processing systems}, pp. 3104-3112. 2014.

[24] H. Min, X. Jiahui, X. Shiguo, and Y. Fuliang. Prediction of chaotic time series based on the recurrent predictor neural
network. {\it IEEE Transactions on Signal Processing},
52:3409–3416, 2004.

[25] C. Xindi, Z. Nian, V. Ganesh K., and W. I. Donald C.
Time series prediction with recurrent neural networks
trained by a hybrid psoˆaA¸Sea algorithm. {\it Neurocomputing}, 70:2342–2353, 2007.

[26] C. Rohitash and Z. Mengjie. Cooperative coevolution of
elman recurrent neural networks for chaotic time series
prediction. {\it Neurocomputing}, 86:116–123, 2012.

[27] He, K., Zhang, X., Ren, S., and Sun, J. Deep residual learning for image recognition. {\it In Proceedings of the IEEE
conference on computer vision and pattern recognition}, pp. 770–778, 2016.

\end{document}